\documentclass{article}

\usepackage{arxiv}
\usepackage[utf8]{inputenc} 
\usepackage[T1]{fontenc}    
\usepackage[authoryear]{natbib}
\usepackage{hyperref}       
\usepackage{url}            
\usepackage{booktabs}       
\usepackage{amsfonts}       
\usepackage{nicefrac}       
\usepackage{microtype}      
\usepackage{lipsum}
\usepackage{graphicx}

\usepackage{wrapfig}
\usepackage{subcaption}
\usepackage{tabularx}
\usepackage{amsmath}
\usepackage{cleveref}
\usepackage{floatrow}
\usepackage{listings}
\usepackage{needspace}
\usepackage{array} 
\usepackage{placeins}
\usepackage[titletoc]{appendix}
\usepackage{float}

\floatstyle{plaintop}
\restylefloat{table}
\usepackage[tableposition=top]{caption}
\usepackage[dvipsnames]{xcolor}

\newcolumntype{Y}{>{\centering\arraybackslash}X}
\newcolumntype{Z}{>{\centering\arraybackslash}p{7cm}}

\definecolor{codegreen}{rgb}{0,0.6,0}
\definecolor{codegray}{rgb}{0.5,0.5,0.5}
\definecolor{codepurple}{rgb}{0.58,0,0.82}
\definecolor{backcolour}{rgb}{0.95,0.95,0.92}

\lstdefinestyle{mystyle}{
    language = Python,
    keywordstyle=\color{blue}\bfseries, 
    stringstyle=\color{Bittersweet}, 
    commentstyle=\color{gray}\itshape, 
    numberstyle=\tiny\color{gray}, 
    identifierstyle=\color{cyan}, 
    basicstyle=\small\ttfamily, 
    breakatwhitespace=false,         
    breaklines=true,                 
    captionpos=b,                    
    keepspaces=true,                 
    numbers=left,                    
    numbersep=5pt,                  
    showspaces=false,                
    showstringspaces=false,
    showtabs=false,                  
    tabsize=2
}

\graphicspath{ {./images/} }

\title{GeoBiked: A Dataset with Geometric Features and Automated Labeling Techniques to Advance Deep Generative Models in Engineering Design.}

\author{
    \hspace{1mm}Phillip Mueller\\ 
    BMW Group \\
    University of Augsburg \\
    \texttt{phillip.mueller@bmw.de}
    \And
    \hspace{1mm}Sebastian Mueller\\ 
    BMW Group \\
    \texttt{sebastian.mm.mueller@bmw.de}
    \AND
    \hspace{1mm}Lars Mikelsons\\ 
    University of Augsburg \\
    \texttt{lars.mikelsonst@uni-a.de}
}

\begin{document}
\maketitle
\begin{abstract}
\textbf{Purpose} – We provide a dataset for advancing Deep Generative Models (DGMs) in engineering design and propose methods to automate data labeling by utilizing large-scale foundation models.

\textbf{Methodology} – GeoBiked is curated to contain 4\,355 bicycle images, annotated with structural and technical features and is used to investigate two automated labeling techniques: 
\begin{itemize}
    \item The utilization of consolidated latent features (Hyperfeatures) from image-generation models to detect geometric correspondences (e.g. the position of the wheel center) in structural images.
    \item The generation of diverse text descriptions for structural images. GPT-4o, a vision-language-model (VLM), is instructed to analyze images and produce diverse descriptions aligned with the system-prompt.
\end{itemize}

\textbf{Findings} – By representing technical images as Diffusion-Hyperfeatures, drawing geometric correspondences between them is possible. The detection accuracy of geometric points in unseen samples is improved by presenting multiple annotated source images. 

GPT-4o has sufficient capabilities to generate accurate descriptions of technical images. Grounding the generation only on images leads to diverse descriptions but causes hallucinations, while grounding it on categorical labels restricts the diversity. Using both as input balances creativity and accuracy.

\textbf{Research implications} – Successfully using Hyperfeatures for geometric correspondence suggests that this approach can be used for general point-detection and annotation tasks in technical images. Labeling such images with text descriptions using VLMs is possible, but dependent on the models detection capabilities, careful prompt-engineering and the selection of input information.

\textbf{Originality} – Applying foundation models in engineering design is largely unexplored. We aim to bridge this gap with a dataset to explore training, finetuning and conditioning DGMs in this field and suggesting approaches to bootstrap foundation models to process technical images.


\textbf{Paper type} Research paper

\end{abstract}

%
%

\section{Introduction}
Rapid advancements in the field of machine learning highlight the pivotal role of high-quality datasets in propelling technological breakthroughs. In Computer Vision, the introduction of high-quality, publicly available datasets has acted as a catalyst to enable researchers to evaluate the performance of diverse methodologies. Datasets like ImageNet \citep{dengImageNetLargescaleHierarchical2009}, CIFAR \citep{krizhevskyLearningMultipleLayers2009} and MNIST \citep{lecunGradientbasedLearningApplied1998} are crucial to level the playing field and set benchmarks that define the state of the art. 

Recent breakthroughs in generative tasks, such as the synthesis of high-quality natural images with Stable Diffusion \citep{rombachHighResolutionImageSynthesis2022}, rely on extensive, high-quality data pools such as LAION \citep{schuhmannLAION5BOpenLargescale2022} for model training. Methodologies for the conditional control of the synthesized content also require data. Training a single adapter to condition images on user-provided sketch-inputs within the ControlNet framework is trained on 500k sketch-image-caption pairs \citep{zhangAddingConditionalControl2023} obtained from an internet database. This amount of data is often out of reach for limited, domain-specific applications of DGM’s in engineering-design. Even approaches that are training-free require benchmark samples to evaluate the effectiveness of the mechanism.

Despite the public availability of general-purpose datasets for various deep learning applications, domain-specific fields like engineering design still face the scarcity of datasets equipped with detailed structural and geometric information. This gap has been highlighted in multiple studies \citep{regenwetterDeepGenerativeModels2022, alamAutomationAugmentationRedefining2024, picardConceptManufacturingEvaluating2023}. It limits the application of Deep Generative Models (DGMs) in engineering design, where the objective extends beyond generating aesthetically pleasing visuals. Fulfillment of fundamental technical feasibility and unambiguous control over the process are prerequisites to develop real-world design concepts.\citep{alamAutomationAugmentationRedefining2024, joskowiczEngineersPerspectivesUse2023}. In such scenarios, precise control over the generative process is necessary. Developing methods for conditioning and controlling the generation requires datasets with design-relevant and interpretable annotations.

In response to these challenges, our work \footnote{Dataset and Code is found under: \url{https://github.com/Phillip-M97/GeoBIKED}} builds upon and extends the foundational efforts of the BIKED project \citep{regenwetterBIKEDDatasetMachine2021}. We streamline and enrich the dataset with  geometric and semantic details aiming to enable engineers and designers to conduct basic experiments on DGMs with structural image data such as model training, finetuning, developing conditioning mechanisms and benchmarking. Our updated dataset includes interpretable features and geometric representations encompassing 12 reference points.

Annotating the required data for generative tasks often is a bottleneck for practical applications. Recognizing this, our work explores the capabilities of utilizing off-the-shelf generative models with capabilities for visual- and language-understanding for automating these tedious manual efforts. We hypothesize that we can leverage the reasoning capabilities and context understanding learned by large-scale foundation models \citep{tangEmergentCorrespondenceImage2023, hwangGroundedIntuitionGPTVision2023, picardConceptManufacturingEvaluating2023, singhAssessingGPT4VStructured2023} to bypass manual data annotation by performing this task with a generative model. In that, we investigate the capabilities of these models to process domain-specific, engineering design image data. 
Specifically, we ask two questions that relate to two relevant data annotation tasks.
\begin{enumerate}
    \item Can we use the learned spatial and semantic understanding of pretrained latent diffusion models (Stable Diffusion \citep{rombachHighResolutionImageSynthesis2022} to detect and annotate geometric correspondences in structural images?
    \item Can state-of-the-art vision-language models (GPT-4o \citep{OpenAIGPT4o}) be used to generate diverse text descriptions of structural image data that accurately describe the technical object and its fine visual details?
\end{enumerate}

The first question corresponds to the task of annotating the structural bicycle images with the geometric features, a task we did by hand for all samples. Analogous to how humans approach such tasks, we aim to provide only a handful of annotated examples. We then utilize the pretrained network by \citep{luoDiffusionHyperfeaturesSearching2023} to consolidate the learned feature representations, extracted from Stable Diffusion into an interpretable tensor, referred to as the Hyperfeature Map. These Hyperfeatures are used to draw correspondences between the geometric features described by the annotated points in the images and geometric features in unseen images. Leveraging the feature representations learned by Stable Diffusion, we automate the task of annotating geometric reference points in the GeoBiked images, showing that a) large-scale diffusion models for image generation can be used to process structural images in engineering design, b) a learned consolidation of latent features can be used to draw correspondences between such technical images, even if it has not been trained on them and c) presenting multiple reference images that show different styles of the design objects leads to higher accuracy in the prediction of the geometrical reference points in unseen images.

For the second question, we aim to generate linguistically diverse text descriptions for the technical images by passing them to GPT-4o \citep{OpenAIGPT4o} as our off-the-shelf vision-language model (VLM). The VLM is instructed to produce descriptions of different lengths and styles. We investigate how the model can be aligned to produce desired descriptions of sufficient diversity without hallucination. Therefore, we evaluate the influence of three different configurations of input information. First, we only input the image for the VLM to describe. Second, we pass the labels annotating the corresponding image. The labels contain information about the style and the technical layout of each bicycle. Finally, we input both previous information types together.

To sum up our main contributions, we present the GeoBiked dataset that contains structural images of bicycles. The images are annotated with interpretable design, technical and relevant geometric features for engineering design applications. The necessary preprocessing steps, the improvements made and the provided semantic and geometric features are discussed in \Cref{sec:GeoBiked}. Based on the GeoBiked dataset, we propose two methods for automated labeling of structural image data utilizing off-the-shelf foundation models, that do not need any training or finetuning in \Cref{sec:Sec3}. First, we show that a learned consolidation of latent image features from Stable Diffusion (Diffusion Hyperfeatures) can be used to accurately predict geometric reference points in unseen images (\Cref{sec:Hyperfeats}). Second, with GPT-4o's vision-language capabilities, we generate diverse text descriptions of controllable style that accurately describe the structural images, balancing the needs for diversity and accuracy (\Cref{sec:TextGen}). \Cref{sec:Sec4} explores potential applications of the dataset, discussing practical use cases and directions for future research. Finally, \Cref{sec:Sec5} concludes the paper by summarizing key findings and contributions.

\section{GeoBiked Dataset}
\label{sec:GeoBiked}
Visual datasets for domain-specific engineering design applications require not only high-quality images but also rich semantic and geometric annotations that make the technical objects in the images interpretable. Such features are essential for enabling Deep Generative Models to perform tasks like conditional control and technically grounded design generation. This section introduces the GeoBiked dataset, which builds on the foundational BIKED dataset \citep{regenwetterBIKEDDatasetMachine2021} by addressing its limitations and enriching it with interpretable design, technical, and geometric features. We begin by describing the baseline dataset, identifying its shortcomings, and detailing the preprocessing and improvements conducted to enhance its quality and usability in \Cref{subsec: Preprocessing}. Subsequently, we outline the dataset’s key features in \Cref{subsec: Features}.

\subsection{Dataset Preprocessing and Improvements}
\label{subsec: Preprocessing}
Our dataset is based on the BIKED-dataset \citep{regenwetterBIKEDDatasetMachine2021}, which originated from the BikeCAD software \citep{CurryBikeCAD2022}, a specialized tool for bicycle design. From an initial collection of 4791 bikes and 23813 descriptive parameters per sample, the authors distilled 4512 bicycles and 1314 parameters. These parameters correlate to the CAD-models and describe every bicycle in detail. This curation process aimed to retain the raw richness of the data, facilitating a broad spectrum of data-driven design applications. Their methodology exemplifies the utility of the dataset through the training of two Variational Autoencoders (VAEs): one dedicated to generating new bike images and another to reconstructing bike parameters, showcasing the versatility of the dataset in supporting innovative design synthesis.

The initial version of the dataset comes with a number of shortcomings, which limit its applicability in DGM-driven engineering design tasks. Each sample is annotated with a total of 1314 parameters that were extracted from the BikeCAD software. Despite large in quantity, these parameters are largely non-interpretable and contain no meaningful information about design, style or structural composition of the bicycles they describe. Having such information is crucial to enable DGMs in engineering and concept design, as they are the basis for conditioning and control modalities of the generatice process \citep{alamAutomationAugmentationRedefining2024, joskowiczEngineersPerspectivesUse2023, regenwetterDeepGenerativeModels2022}. Furthermore, the dataset contains a number of infeasible designs and unrealistic samples. It lacks a uniform scaling of the objects within the image resolution, preventing direct geometric correspondences between the images. 

\begin{figure}[h]
\centering
\includegraphics[width=\linewidth]{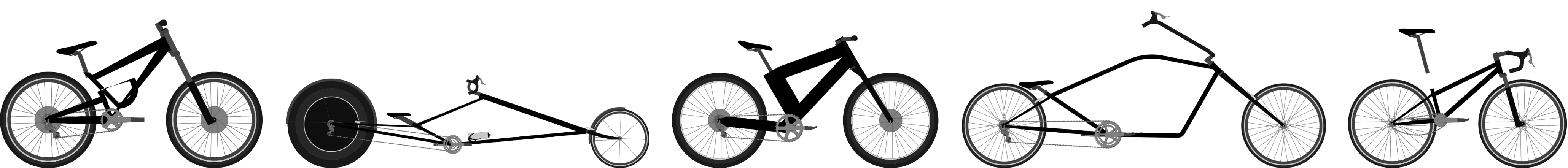}
\caption{Unrealistic samples from the original BIKED dataset \citep{regenwetterBIKEDDatasetMachine2021}.}
\label{fig:Unrealistic}
\end{figure}

To address the aforementioned issues, we conducted a thorough curation process, improving the dataset's quality and usability in engineering design. The first curation step in the derivation of GeoBiked from BIKED is a visual inspection of the 4512 provided bike images for faulty and out-of-distribution samples. We categorize samples as faulty if their geometric integrity is not ensured or the frame design is visibly unrealistic, see \Cref{fig:Unrealistic}. We also remove the ten geometrically largest samples from the dataset as they are significantly out-of-distribution, therefore reducing the variance of the bike sizes and the geometric characteristics by more than 30\% (\Cref{fig:Variance_Reduction}).

The next step is the centering and geometric normalization of the images. In the original dataset, the images are not normalized to the same scale. Using the provided information about the scaling factor in the BIKED dataset, we ensure that all bikes are scaled based on the same scaling factor, solidifying geometric consistency. Furthermore, we maximize the size of the bikes within the image resolution by precisely fitting the largest sample to the image resolution in length (x-dimension), which leads to one pixel equaling 10.19 mm in $256 \times 256$ images and 1.27 mm in $2048 \times 2048$ images.

To add geometric reference points to the dataset, we first define characteristic points and intersections in the bike geometry, aiming to allow the representation of each sample solely by a combination of these points. We describe the points in more detail in Section \ref{sec:Geometric_Features}. Of the 12 geometric points, we derive six from the provided parameter set in the original dataset. The remaining parameters have to be defined manually. The coordinate-values of the geometric points are stored in millimeters for simplicity. They can be translated into pixel-values with the previously discussed scaling for both image resolutions.
We filter and modify the initial set of descriptive parameters for features with semantic, geometric, or technical relevance, keeping a total of nine. We discuss the provided features in the next section in more detail.

\subsection{Dataset Features}
\label{subsec: Features}
\subsubsection{Design Features}
We categorize the samples in the dataset into 19 different bicycle styles. This categorization is adapted from the original BIKED dataset. The style distribution is shown in Figure \Cref{tab:Bike_Style}. In the dataset, Road-Bikes are the most common style, followed by Mountain-Bikes and Track-Bikes.

We further add the diameters of the front and the rear wheel to the dataset, which are essential for the overall bicycle composition and geometry and can therefore be used as technical or design parameters. Furthermore, we provide the possibility of categorizing the samples by their Rim-style. The Rim-style has a significant impact on the overall design and appearance of the bicycle. The dataset distinguishes between spoked rims, tri-spoked rims and disked rims. Front-and rear wheel rim styles are separate categories, as several samples have a combination of two different rims. We note that spoked rims are by far the most common category in the dataset, making up 93.5\% of samples for the front wheel and 92.9\% for the rear wheel. 

Another design-related feature that we provide in our dataset is the fork type. We distinguish between rigid forks, suspension forks and single-sided forks. Rigid forks make up 77.1\% of samples while suspension forks make up 20.2\% and single-sided forks 2.6\%. 
The final set of design features are the bottles on the seat-tube and the down-tube. We added information about their presence as Boolean-values to the dataset. 

\begin{table}[h]
\captionof{table}{Quantity Distribution of Bike Style (left), tube diameters (middle) and frame sizes (right).}
\centering
\begin{minipage}{0.58\textwidth}
    \scriptsize
    \centering
    \begin{tabularx}{\linewidth}{cccccc}
    \toprule
    \multicolumn{6}{c}{\textbf{Bike Styles}} \\
    \midrule
    {Category} & {Quantity} & {Category} & {Quantity} & {Category} & {Quantity} \\
    \midrule
    ROAD & 1802 &  CYCLOCROSS & 150 &  CRUISER & 37 \\
    MTB & 600 & POLO & 128 & HYBRID & 34 \\
    TRACK & 448 & BMX & 85 & TRIALS & 28 \\
    OTHER & 294 & TIMETRIAL & 75 & GRAVEL & 19 \\
    DIRT-JUMP & 292 & COMMUTER & 74 & CARGO & 12 \\
    TOURING & 197 &  CITY & 70 & CHILDRENS & 10\\
    \bottomrule
    \end{tabularx}
    \label{tab:Bike_Style}
\end{minipage}
\hfill
\begin{minipage}{0.19\textwidth}
    \scriptsize
    \centering
    \begin{tabularx}{\linewidth}{cc}
    \toprule
    \multicolumn{2}{c}{\textbf{Tube Diameters}} \\
    \midrule
    {Category} & {Quantity}\\
    \midrule
    Mini & 946\\
    Lite & 474\\
    Standard & 1995\\
    Reinforced & 651\\
    Extreme & 289\\
    \bottomrule
     & \\
    \end{tabularx}
    \label{tab:Tube_Size}
\end{minipage}
\hfill
\begin{minipage}{0.19\textwidth}
    \scriptsize
    \centering
    \begin{tabularx}{\linewidth}{cc}
    \toprule
    \multicolumn{2}{c}{\textbf{Frame Sizes}} \\
    \midrule
    {Category} & {Quantity}\\
    \midrule
    XS & 529\\
    S & 175\\
    M & 426\\
    L & 1800\\
    XL & 1425\\
    \bottomrule
    & \\
    \end{tabularx}
    \label{tab:Frame_Size}
\end{minipage}
\end{table}

\subsubsection{Technical Features}
To provide technical features, we categorize the samples by their tube-sizes, their frame-sizes and the number of teeth on the chainring. For the tube-sizes, each bicycle consists of four major tubes. To categorize the samples in a meaningful way, we calculate the average tube diameter for each of the four tubes over the entire dataset. The average seat-tube-diameter is 31.5mm, the average down-tube-diameter is 35.5mm, the average head-tube is 42.9mm in diameter and the top-tube measures at 32.0mm. The classification is carried out by counting how many diameters of a given sample are greater than the average. If all diameters are smaller than their class-average ($n_{TS}=0$), the sample is considered to have a “minimal” tube-size. For $n_{TS}=1$, the tube-size is considered “lite” and “standard” for $n_{TS}=2$. If $n_{TS}=3$ the tubes are “reinforced” and for $n_{TS}=4$ they are categorized as “extreme”. The frequency of each category is visualized in Figure \Cref{tab:Tube_Size}.

In addition to the tube sizes, we provide information about the bicycles’ frame sizes, as this is a common metric for categorization. The most common way to determine the frame size is to measure the length of the seat-tube between the bottom bracket and the top edge of the tube \citep{normanHowMeasureBike2024}. We consider all samples with a seat-tube length smaller than 360 mm to be “XS”. For seat-tube lengths between 360mm and 420mm, the samples are considered ``S'' and for lengths between 420mm and 480mm they are considered size “M”. A bicycle is of frame-size “L” if the seat-tube length is between 480mm and 540mm and “XL” if it is longer than 540mm. The distribution of frame sizes is  shown in Figure \Cref{tab:Frame_Size}.

\subsubsection{Geometric Features}
\label{sec:Geometric_Features}
\begin{figure}[h]
\centering
\includegraphics[width=\linewidth]{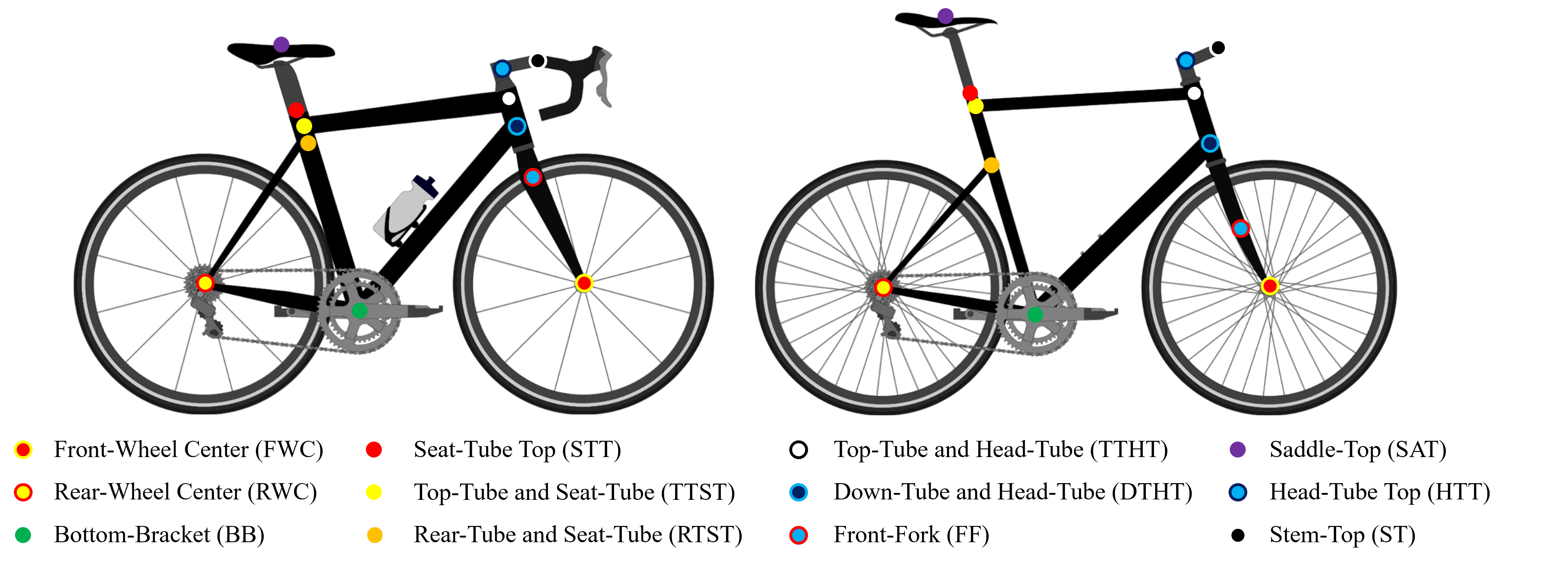}
\caption{Geometric layout described by the geometric reference points.}
\label{fig:Geo_Points}
\end{figure}

With the selection of the 12 reference points, we aim to capture bicycles in all styles and sizes and characterize them by their geometrical layout. The final selection of 12 geometric points is shown in Figure \Cref{fig:Geo_Points}. We select the center points of the rear wheel (RWC) and the front wheel (FWC) for obvious reasons as they define the wheelbase of the bicycle. The point “BB” marks the center of the intersection of the seat-tube and the chain-stay and therefore the center of the bottom bracket. The head-tube-top (HTT) marks the upper end of the tube that connects the handlebars to the fork, given that this is where the stem intersects with the fork-tube.  The stem-top (ST) marks the end of the stem and is of significant influence for the reach of the bicycle frame \citep{normanHowMeasureBike2024}. We define another point on the front-fork (FF), characterizing its shape and potential bends or angles. The seat-tube-top (STT) marks the upper end of the seat-tube and the saddle-top (SAT) describes the highest point of the saddle. 

In addition to the points already mentioned, we define every intersection of the tubes that make up the bicycle frame. Namely, the intersections of top-tube and seat-tube (TTST), top-tube and head tube (TTHT), down-tube and head tube (DTHT), rear-tube and seat-tube (RTST). All points are defined by their x- and y-coordinate values, given in millimeters. The values are relative to the rear-wheel-center, which we defined as the center of the local coordinate system. Since the position of RWC in the image varies from sample to sample, we provide two additional values, $x_{zero}$ and $y_{zero}$, that describe the distance of RWC to the bottom-left corner of the image. This allows for accurate localization of the geometric points within an image as well as comparisons of the geometric layout and bicycle sizes.

\section{Automated Dataset Annotation Techniques}
\label{sec:Sec3}
Automating the annotation of domain-specific datasets is a critical step toward enabling scalable and efficient applications of Deep Generative Models in engineering design. Manual annotation is time-intensive and prone to inconsistencies and biases, particularly for tasks requiring precise geometric labeling or diverse textual descriptions. To address these challenges, this section explores two complementary methods for automating dataset annotation. First, we focus on structural feature detection to annotate technical images with geometrical reference points. We utilize the spatial and semantic understanding of pretrained diffusion models in the form of Diffusion Hyperfeatures \citep{luoDiffusionHyperfeaturesSearching2023} to reduce the reliance on manual labeling, thereby enabling more scalable dataset creation.
The second method employs the Vision-Language Model (VLM) GPT-4o \citep{OpenAIGPT4o}, to generate diverse and accurate text descriptions for structural images. These descriptions can facilitate text-based conditioning of DGMs, a critical requirement for many design applications.

The remainder of this section is organized as follows. First, we present a combined discussion of related work about automated dataset annotation techniques. This provides context for the two annotation methods we propose. Following this, we provide a detailed discussion of both annotation methods: geometric feature detection using Diffusion Hyperfeatures in \Cref{sec:Hyperfeats} and text description generation with VLMs in \Cref{sec:TextGen}. Each subsection covers the respective methodology, experimental setup, and a discussion of results.

\subsection{Related Work}
\subsubsection{Geometric Feature Detection}
Automatically annotating images with geometric reference points poses the task of detecting geometric features. Traditional keypoint detection methods, such as Mask R-CNN \citep{he2018maskrcnn} and Vision Transformers (ViTs) \citep{dosovitskiy2020vit}, rely on large quantities of annotated data for supervised learning. These approaches are typically tailored to detect a predefined, fixed set of keypoints and lack the flexibility needed for diverse, domain-specific tasks.

A number of works have recently addressed the idea of utilizing and transferring learned latent representations from large-scale models for downstream tasks \citep{GoodwinObjectPoseEstimation2022, oquabDINOv2LearningRobust2024, tumanyanPlugandPlayDiffusionFeatures2022}. From various studies and applications, we know that those representations carry information about the underlying structure and composition of images \citep{fanNoiseSchedulingGenerating2023, sabour2024align}. The possibility to draw semantic correspondences with these deep features has also been proven \citep{amirDeepvitfeatures2022, CaronDINO2021, oquabDINOv2LearningRobust2024}. 

Recently, \citep{luoDiffusionHyperfeaturesSearching2023} have made significant progress in this domain by consolidating the latent feature maps from SD, extracted over multiple layers and timesteps, into an interpretable, per-pixel descriptor. The so-called Hyperfeature map allows to draw semantic correspondences between two images by comparing their respective Hyperfeature maps. In their work, they train an aggregation network to consolidate an image into its Hyperfeature Map. This is done by caching the intermediate feature maps obtained by either generating a synthetic image or inverting an existing image through multiple diffusion timesteps. Each extracted feature map is upsampled to a standard resolution and passed through a bottleneck layer for standardization. Subsequently, mixing weights are learned that identify the most significant features. For the task of semantic correspondence, the cosine similarity is computed between the flattened descriptor maps of image pairs that are labeled with corresponding keypoints. The aggregation network is trained by minimizing a symmetric cross-entropy loss similar to CLIP \citep{radfordLearningTransferableVisual2021} between the predicted and ground truth keypoints.

\subsubsection{Generation of Text Descriptions}
Generating text descriptions fundamentally is an image captioning task. Previous works address this topic with a variety of approaches, utilizing CLIP-embeddings \citep{mokady2021clipcap} or vision-language transformers \citep{zhou2019vlp, liVisualBERTSimplePerformant2019}. The development of GPT-4(V) is a major improvement as it possesses vast context understanding and broad general reasoning capabilities to accurately caption contents of an image \citep{singhAssessingGPT4VStructured2023, hwangGroundedIntuitionGPTVision2023}. Besides the multi-modal variations of GPT-4, other vision-language models like LLaVA \citep{liuVisualInstructionTuning2023, liuImprovedBaselinesVisual2024} are available, but are outperformed on the VisIT-Benchmark for VLMs \citep{bitton2023visitbench} by GPT-4(V) in terms of model performance across a diverse set of instruction-following tasks.

The comprehensive study by \citep{picardConceptManufacturingEvaluating2023} investigates the feasibility of including GPT-4(V) in an engineering design process and thereby underlines our assumptions that the capabilities of the model are sufficient for domain-specific captioning tasks on technical images.

In terms of using language models for generating synthetic descriptions of images, a few works are to be named. Cosmopedia \citep{benallal2024cosmopedia} is a dataset of synthetic textbooks, blogposts, stories, posts and WikiHow articles generated by Mixtral-8x7B-Instruct \citep{Mistral7b}.The dataset contains over 30 million files and 25 billion tokens. BLIP-2 employs a two-stage pre-training approach to enhance image-grounded text generation. By using transformers for both image and text processing, BLIP-2 generates accurate and contextually relevant descriptions from images. It demonstrates the ability to perform zero-shot image captioning effectively, making it versatile for various applications \citep{Li2023BLIP}. LAVIS is a whole suite for developing multi-modal models, allowing to rapidly employ and benchmark models for different tasks, image captioning being one of them \citep{li-etal-2023-lavis}. 
DreamSync \citep{sun2024dreamsync} employs VLMs to automate the selection of high-quality image-text pairs to finetune a text-to-image model. \citep{garg2024imageinwords} employ a VLM together with human annotators to create a dataset of hyper-detailed, synthetic image descriptions.

\subsection{Geometric Feature Detection with Diffusion Hyperfeatures}
\label{sec:Hyperfeats}
Adding structural or geometric information to technical images is fundamentally important for the application of DGMs in engineering design. Despite the recent explosion in publications about visual DGMs, there still is a shortage of datasets with domain-specific modalities that enable conditional control over the generative process. However, this control is a key requirement for the successful application of DGMs for technical and design tasks \citep{alamAutomationAugmentationRedefining2024, MuellerExploringthePotentials2024}. In GeoBiked, we provide information about the geometric layout of each image sample. This might be used to investigate geometry-aware image generation or conditioning a model on a geometric layout. In annotating the geometric points largely by hand, we come to the conclusion that this time-intensive manual task poses a significant barrier for domain-specific applications. We therefore aim to utilize the spatial and semantic understanding inherited by pretrained, large-scale diffusion models for image generation (Stable Diffusion) and automate the annotation of image data with geometric reference points \citep{poStateArtDiffusion2023, tangEmergentCorrespondenceImage2023, luoDiffusionHyperfeaturesSearching2023}. 

\subsubsection{Method}
We propose to utilize the aggregation network pretrained for semantic correspondence to draw geometric correspondences in the GeoBiked images. In their work, \citep{luoDiffusionHyperfeaturesSearching2023} already show that their approach outperforms existing alternatives for matching keypoints in natural images. We apply their methodology for our dataset and extend it to be able to handle multiple annotated source images. We draw inspiration from the manual labeling process, where a human annotator is shown a small number of annotated reference images that show the relevant features and subsequently identifies them in unseen images showing similar concepts. The possibility to base the prediction of the geometric reference points on multiple source images is introduced to make the prediction more reliable given the variety of bicycle geometries and styles in the dataset.

The source images $i \in I_s$ are annotated with their corresponding geometric points $p_{ik}$. It is worth noting that while the points can be chosen freely, they have to be consistent inbetween the source images, meaning that they mark the same geometric characteristics. Using the pretrained aggregation network, the Hyperfeature map ${H_{s_i} \in \mathbb{R}^{C \times 64 \times 64}}$ is computed for each source image, where $C$ is the channel dimension. Depending on the resolution of the source images ($H \times W$), the Hyperfeatures are interpolated back to the original image size to obtain $H'_{s_i} \in \mathbb{R}^{C \times H \times W}$. They are subsequently flattened and normalized to form $F_{s_i} \in \mathbb{R}^{(H \cdot W)\times C}$. Per source image, each labeled point $k \in p_{ik}$ is translated into its index $idx_{ik}$ to extract the corresponding Hyperfeatures $V_{s_i}$ from the flattened map $F_{s_i}$: 
\begin{equation}
    \label{Extract_Source_Hyperfeats}
    V_{s_i} = F_{s_i}[:,idx_{ik},:] \text{ with } V_{s_i} \in \mathbb{R}^{N \times C}.
\end{equation}

When processing an entire dataset, the unlabeled images to be annotated are processed one after another. For each target image $I_t$, we compute its Hyperfeature map $H_t \in \mathbb{R}^{C \times 64 \times 64}$, interpolate it to the original image size $H'_{t} \in \mathbb{R}^{C \times H \times W}$ and also flatten and normalize it for $F_t \in \mathbb{R}^{(H \cdot W) \times C}$.
Given the Hyperfeature representations of the source points $V_{s_i}$ and the target image $F_t$, we now can compute the similarity matrix:
\begin{equation}
    \label{Sims}
    S_i = V_{s_i} \times F^{T}_t  \text{ with } S_i \in \mathbb{R}^{N \times (H \cdot W)}.
\end{equation}

We obtain one similarity matrix of size $N \times (H \cdot W)$ per combination of source and target image, where $N$ is the number of points. The similarity matrix contains the cosine similarities between the Hyperfeatures of the annotated source points and the Hyperfeatures of the entire target image. Now we extract the maximum cosine similarities per row in $S_i$ to obtain $v_i$, which describes the Hyperfeatures in the target image that have the highest correspondence to the Hyperfeatures of source image $i$ with respect to the points. Since we are processing multiple source images, we can concatenate the similarity vectors $\mathbf{v}_i$ along the y-axis and now extract the row-wise maximum similarity:
\begin{equation}
    \label{Max_Sims}
    \mathbf{v}_{max}[k] = \max_{i} (\mathbf{v}_i[k])
\end{equation}
This increases flexibility in the prediction of the geometric points as per-point, the source-target combination with the highest correspondence is selected. In simple terms, this allows us to predict the positions of the geometric reference points in the target images using the information from all the source images. We always chose the position of the point where the correspondence between the Hyperfeatures of the source and target image is highest.

\subsubsection{Experiments}
In our experiments, we aim to verify the hypothesis that we can in fact use the diffusion Hyperfeatures to detect the geometric features in unseen structural images. Furthermore, we want to find the optimale selection and quantity of annotated source images to accurately label the diverse bicycle images in the dataset. For the evaluation, we select a subset of 150 diverse samples from our dataset. All calculations are conducted on an NVIDIA RTX A4500 with 20GB of VRAM. The results are summarized in \Cref{tab:Hyperfeats_Experiment} as well as in \Cref{fig:Error_over_Quant}. For evaluation, we use the pixel-wise MAE and MSE between the predicted location of the point and the ground truth location that we annotated by hand. We average the errors over all annotated points per image to gain insights on how well the entire geometric layout is captured and predicted.

\begin{table}[h]
\caption{Experiment results for eometric feature detection in the GeoBiked subset. The accuracy of the geometric point detection depends on both the selection and quantity of the annotated source images. For each source image quantity, multiple bicycle types were tested (e.g., road bikes, BMX, mountain bikes) to evaluate the variability of results. The table reports the best-performing result for each tested source image quantity. The metrics are computed as a pixel-wise distance between prediction and ground truth, averaged over all 12 geometric points. The duration measures the processing time for the subset of 150 samples.}

\centering
\begin{tabularx}{\columnwidth}{YYYY}
\toprule
{Source Image Quantity} & {MAE $\downarrow$} & {MSE $\downarrow$} & {Duration (min.)}\\
\midrule
\textit{1} & 2.429 & 33.158 & 21.76\\
\textit{2} & 2.063 & 13.836 & 22.11\\
\textit{3} & 1.989 & \textbf{11.367} & 24.97\\
\textit{4} & 1.943 & 12.251 & 27.76\\
\textit{5} & 2.025 & 12.256 & 28.15\\
\textit{6} & \textbf{1.845} & 11.733 & 26.82\\
\textit{7} & 2.028 & 14.204 & 26.12\\
\textit{8} & 2.001 & 13.928 & 30.74\\
\bottomrule
\end{tabularx}
\label{tab:Hyperfeats_Experiment}
\end{table}

\textbf{Single Source Image.} When providing only one source image, we observe that the accuracy of the point-location prediction is heavily dependant on the type of source image that is provided as reference (see \Cref{tab:SingleSourceImg} and \Cref{fig:Error_Patterns_SSI}). We selected various bicycle styles as reference images and observed that an average geometry, which captures a wide selection of bicycle frames, leads to good prediction accuracy as the MAE is below 3 pixels. However, with this kind of source image, the method is not able to draw accurate correspondences to bicycles with a significantly different geometric layout. This is evident in the large MSE. When samples are chosen as source images that are not in the middle of the geometric distribution, the prediction accuracy deteriorates significantly, as seen in \Cref{tab:SingleSourceImg}. For the source image with the best prediction accuracy, we still observe typical error patterns. Tube intersections are often annotated inaccurately as well as the saddle top and points around the handlebars. For the annotation of outlier samples, only the wheel centers are captured reliably.

The observed ambiguities most likely originate from the structure of the bicycles in the images. Due to them being plain grayscale structures on a white background, areas with tube intersections look very similar. A single source image does not capture the different options of the bicycle layout with respect to the saddle position, stem and handlebars and tube intersections. For example, if multiple tube intersections fall into the same position in the source image, but are in different positions in the target image, they will not be detected with high precision.

\begin{figure}[h]
\centering
\includegraphics[width = \linewidth]{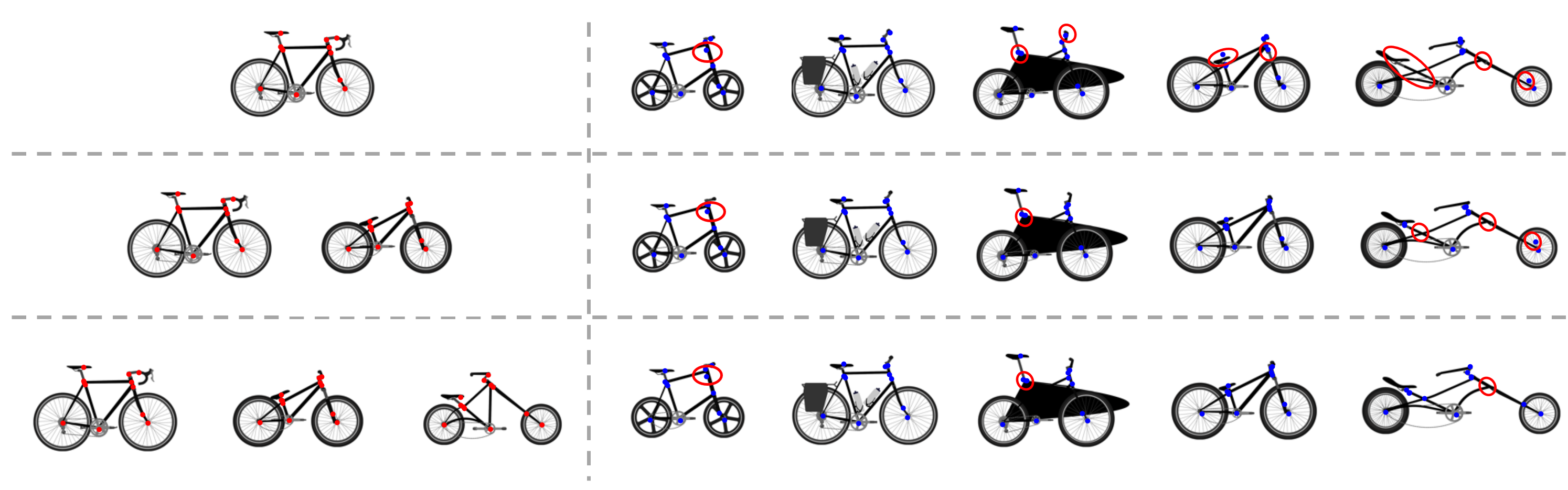}
\caption{Qualitative Comparison of error patterns using one, two and three source images respectively. Images in the left columns (red marks) are source images. Images in the right column (blue marks) are target images. Most of the inaccuracies disappear when using three source images compared to a single one. Some uncertainty remains when uncommon samples are processed (see middle and last column). Red circles in the target images mark areas of inaccuracy in the point prediction. Best viewed when zoomed in.}
\label{fig:Error_Patterns_compare}
\end{figure}

\textbf{Multiple Source Images.} Comparing an unlabeled target image with multiple source images for geometric correspondence noticeably improves the accuracy of the point prediction. By just using a second source image showing a different type of bicycle, we reduce the MSE by a factor of 2.4. The variety of layouts in the dataset gets captured much more reliable. Since we can precompute the Hyperfeatures for the source images once and then use them for processing the entire dataset, the computational overhead is insignificant. For two source images, we test a variety of combinations. We observe that using the sample that performs best when used as the only input together with a sample showing a different, but common, style leads to the best accuracy. Typical error patterns observed in the previous section are largely eliminated (see \Cref{fig:Error_Patterns_compare} and \Cref{fig:Error_Patterns_2SI}). Nevertheless, for uncommon styles the prediction is still inaccurate.

Based on the observations with one and two source images, we add a third source image showing such sample. With that, we are able to further reduce the MSE by about 18\%. In our experiments, we observe that increasing the number of source images beyond three does not have a significant impact on the prediction accuracy (see \Cref{tab:Hyperfeats_Experiment} and \Cref{fig:Error_over_Quant}). We therefore propose that, in the case of the GeoBiked dataset, using three source images presents a good balance of manual annotation effort and prediction accuracy of the automated process. Using the three source images shown in \Cref{fig:Error_Patterns_compare}, we also process the entire GeoBiked dataset, achieving an accuracy of $MAE = 1.837$ and $MSE = 14.009$ in a processing duration of 11.33 hours.

\subsubsection{Implications}
For engineering design images with fine-grained details, the provided source images have to capture the variety of structures in the target image space as much as possible. It is important to select the references so that they show different layouts of the object geometry. In our case, this correlates to bicycles with overlapping and non-overlapping tube intersections. When processing a diverse dataset, outliers and uncommon samples require specific attention. They need to be annotated by and or at least evaluated for sufficient accuracy. When these limitations are regarded for, pretrained diffusion models can be used to automate the process of data annotation. Finetuning of the aggregation network might be necessary for some applications where the off-the-shelf accuracy is not sufficient.

Our experiments show that representing images through their diffusion Hyperfeatures generally allows us to process domain-specific, structural image data. Even though the diffusion model was not specifically trained on such data, the learned semantic correspondences and spatial understanding can be transferred and applied to engineering design tasks. Efficiently using the inherited capabilities however is not straightforward, as the feature representations extracted from the diffusion model need to be further processed. In this case, training an encoder-like model is required to draw semantic correspondences. Nevertheless, compared to a full-scale training or finetuning of a diffusion model this is relatively cheap. The aggregation network was trained on a single NVIDIA RTX TITAN 24GB GPU \citep{luoDiffusionHyperfeaturesSearching2023}. Computing the Hyperfeatures of an image and using them for downstream applications can also be considered efficient as it is possible on a consumer-grade GPU.

In an outlook, we propose that Diffusion Hyperfeatures of images can also be used for other engineering design applications in image generation and modification. A future research direction can be to utilize them in order to improve object consistency in generative processes by comparing the Hyperfeature representations of the generated object with a learned distribution or ground-truth example. In a different work building up upon Diffusion Hyperfeatures, the authors address this task in broad terms \citep{luo2024readoutguidance}. 


\subsection{Automatic Generation of Text Descriptions with Vision-Language Models}
\label{sec:TextGen}
The possibility of text-based conditioning is a key factor of the recent success in image generation as it provides an intuitive modality to control the generated content. Numerous works discuss this topic. Most notably \citep{pmlr-v139-ramesh21a, rombachHighResolutionImageSynthesis2022, sahariaPhotorealisticTexttoImageDiffusion2022, ruiz2022dreambooth, midjourneyinc.Midjourney2023} in image generation and \citep{sahariaPaletteImagetoImageDiffusion2022, brooks2022instructpix2pix, Yang_2023_CVPR, tumanyanPlugandPlayDiffusionFeatures2022} in image-to-image editing. It provides a way to dynamically adjust the amount of information passed to the model. For engineering design applications, the DGM can be trained to adhere to many design constraints and requirements as well as fill in the blanks for short, high-level descriptions. Training such models and corresponding conditioning mechanisms requires pairs of images and diverse text-descriptions. For better generalisation, the text-descriptions have to contain varying amounts of information.

Labeling domain-specific images with diverse text-prompts by hand is generally unfeasible. In addition to the high manual effort on a repetitive task, the descriptions would follow the bias of the annotators and be limited in their diversity and creativity. When performing repetitive tasks like this, human annotators are prone to produce text descriptions with low linguistic variety, frequently reusing similar phrasing and structures across multiple samples. Therefore, we provide a recipe on how to bootstrap the capabilities of large-scale (vision-) language models to automate this task. We employ GPT-4o \citep{OpenAIGPT4o}, which build upon GPT-4 \citep{OpenAIGPT4Tech} and GPT-4(Vision) \citep{openaiGPT4VIsionSystem2023} with the additional benefit of providing noticeably faster inference. We use this model because of its state-of-the-art vision-language understanding and reasoning and its easy accessibility through the API. In addition to providing a method for automatic labeling, this task provides insights on the capabilities of GPT-4o for context understanding and reasoning in technical images with fine-grained details.

\subsubsection{Method}
We access GPT-4o through the API and prompt it to generate descriptions of the GeoBiked bicycle images. To obtain diverse descriptions of the images, we prompt GPT-4o to adhere to specific description characteristics. The system-prompt passed to GPT-4o specifically outlines the format of the generated description. The mask for the system-prompt is provided in \Cref{sec:Code}.
\begin{itemize}
    \item \textbf{Length:} We distinguish between short (5-10 words), medium (10-20 words) and long (20-40 words) descriptions.
    \item \textbf{Character:} The model is prompted to generate descriptions with either a technical character or a casual description.
    \item \textbf{Style:} The descriptions have to be either in style of a marketing message or of a prompt for a text-to-image model (Midjourney).
\end{itemize}

We generate text-descriptions with three different data sources that are provided to GPT-4o. For the first one, only the bicycle image is provided. It is wrapped in the system-prompt after the general task description and the definition of the required description characteristics. The general task instructs to model to analyse the input information and create a description for it while adhering to the formulated description characteristics. The second data source are labels in the form of technical categories describing the bike. No image is used for this configuration. From GeoBiked, we use bicycle style, rim styles, fork type and whether there are bottles on the down- or seattubes. In the third configuration, we pass the image together with the technical categories. 

The motivation for selecting the three input information combinations is grounded on several ideas. By providing only the image, we generally test if the VLM has sufficient context understanding and visual detection capabilities to capture the details of the presented bicycle images. The label-only configuration is used for comparison and for verification that the increased information density of an image compared to five categorical features actually leads to more diverse and accurate descriptions. With the third setup, we test if both information inputs used together hold an advantage over any combination on its own.

\subsubsection{Experiments}
\label{subsec:gpt_metrics}
First and foremost, we aim to verify if GPT-4o is able to generate creative and diverse descriptions of the samples from GeoBiked, given that the images are of a technical character and contain fine-grained visual details. We further analyze if the generated descriptions follow the required length, character and style. By also passing ground-truth labels to the model, we evaluate the possibility of generating diverse, but hallucination-free descriptions as hallucination is a well-known issue with LLMs \citep{Huang2023ASO}. 

In terms of evaluation metrics, we measure diversity by counting the number of unique outputs generated with different description requirements. This metric provides insights on the repetitiveness of the generated descriptions and therefore indirectly measures how well the VLM can capture different inputs and create unique descriptions from them.
Additionally, we compute the Levenshtein-distance between unique outputs as a measurement for the difference between two linguistic sequences \citep{Haldar2011LevenshteinDT}. A higher average Levenshtein-distance indicates more diverse text-descriptions. We use the implementation by \citep{Bachmann2024}. 

To evaluate the accuracy of the generated text-description with the ground-truth labels, we again use GPT-4o. In this setting, the model is utilized as a classifier. We instruct it to infer the categorical labels from the text description and then compare the extracted labels from the text-description to the ground-truth labels and count the inconsistencies. The code is provided in the linked repository. We acknowledge that this approach adds a degree of uncertainty due to the classifier having to infer the labels from the text-description. We assume that the error injected by this process is consistent and we therefore can still derive correlations between the system-prompt configuration, the provided input information and the accuracy of the generated text-prompt.  

We conduct our experiments on the entire GeoBiked dataset, containing all 4355 samples. We use the images with $2048 \times 2048$ resolution. For better comparison, we only use one thread, calling the API one time after another. In terms of hardware, for CPU we use 12th Gen Intel Core i7-12850HX 2.10 GHz. Generating the descriptions for each of the three input configurations takes approximately 15 hours.

\begin{figure}[h]
\includegraphics[width=\linewidth]{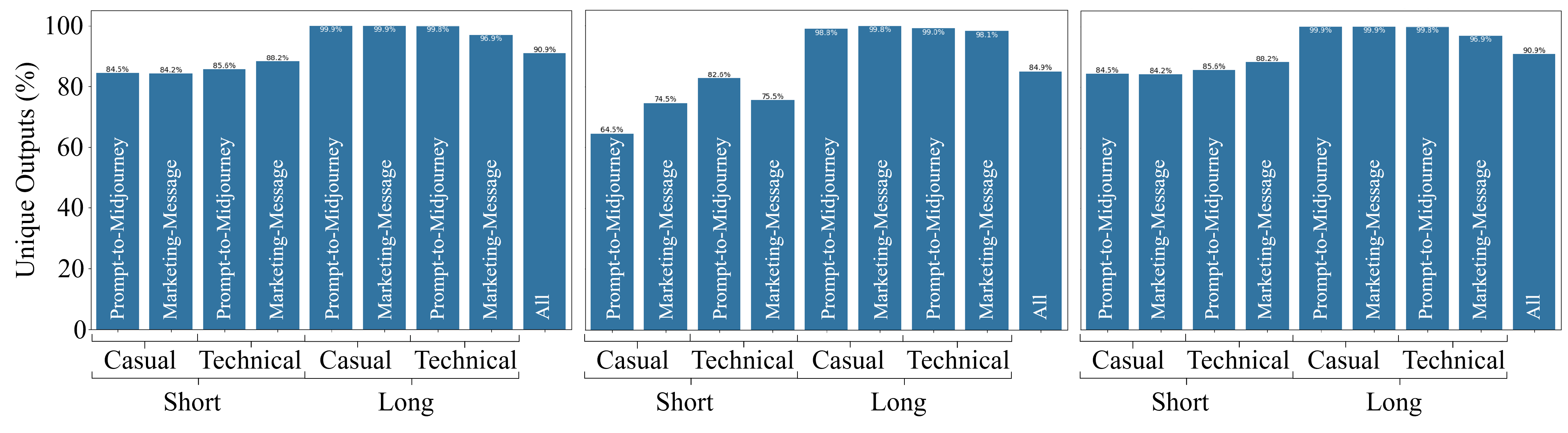} 
\caption{Percentage of unique descriptions generated by GPT-4o, compared over different configurations of description characteristics. For the configuration of description characteristics, the length and character are denoted on the x-axis while the style is given for each bar of the plot. Left: Image only, Middle: Ground-Truth Label, Right: Image and Ground-Truth Label. Best viewed zoomed in.}
\label{fig:Unique}
\end{figure}

Our results for description uniqueness are visualized in \Cref{fig:Unique}. When only the image is passed to GPT-4o, we observe that for short descriptions, between $84.5\%$ and $88.2\%$ of the outputs are unique. Short descriptions therefore tend to be repetitive in some cases (see \Cref{fig:Unique}). The long descriptions are almost entirely unique (between $96.9\%$ and $99.9\%$). On average, we obtain $90.9\%$ unique descriptions. This is also the case when an image together with the ground-truth label is passed (\Cref{fig:Unique}). When the ground-truth labels are used as sole input, the uniqueness of the generated descriptions is significantly reduced for the short ones (\Cref{fig:Unique}). Even if the temperature, a parameter for randomness and therefore diversity of outputs, is set to a high value of 1, the label grounding  restricts the model so much that the uniqueness is noticeably reduced. 

\begin{figure}[h]
\centering
\includegraphics[width = 0.9\linewidth]{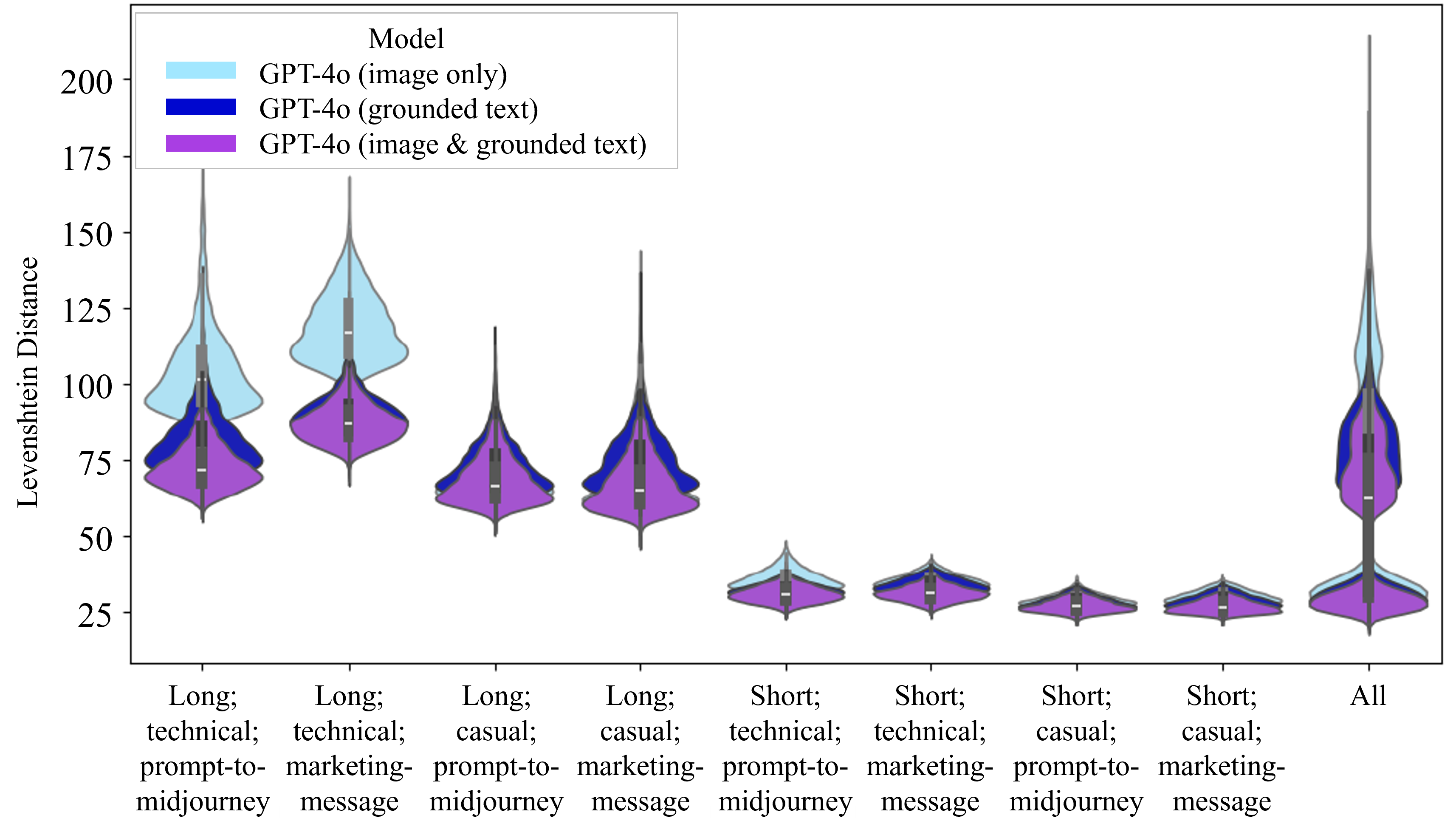}
\caption{Levenshtein distances between unique values for all three configurations of inputs.}
\label{fig:levenshtein_comp}
\end{figure}

The analysis of the Levenshtein distance as a measure of diversity and creativity supports the previous findings. Providing only an image leads to a greater Levenshtein distance among unique text-descriptions, as shown in the violin plot (\Cref{fig:levenshtein_comp}). For image-grounded text-descriptions, the average Levenshtein distance is significantly higher than for label-grounded descriptions, especially for long, technical ones. Grounding the generation on image and labels seems to also restrict the diversity of the unique outputs. As this configuration of input information generally produces more unique outputs than label-only grounding, we assume that providing the image leads to diverse descriptions while the label streamlines the descriptions in terms of format and sentence structure, hence the lower Levenshtein distance between unique descriptions. 

For short descriptions, the distances of the unique outputs are again greater for image-only grounding, although the difference is smaller. A fact to keep in mind here is that image-grounding produces far more unique outputs in total.

When manually evaluating the generated text descriptions, we find that the ones generated on image grounding are more creative overall. For the label-grounded descriptions, we observe a pattern that they tend to be a concatenation of the ground-truth labels wrapped with some filling text. Furthermore, some design characteristics of bicycles are not captured by the ground-truth labels and therefore not regarded for in the generated description. Examples of this can be found in \Cref{tab:Example_Descriptions}. The bicycle in column four is correctly identified and described as vintage or retro when using the image for grounding, while the label-grounded descriptions do not mention anything about the extraordinary design. The same is true for the example given in column five, where the pannier, that is clearly visible in the image, is correctly mentioned for both image-grounded descriptions, but not for the label-based descriptions.

Besides diversity in generated outputs, we measure the accuracy of the generated descriptions to verify that they fit the image they aim to describe. The accuracy is calculated by instructing GPT-4o as a classifier and count the quantity of features that are described inaccurately, as described in \Cref{subsec:gpt_metrics}. For the results in \Cref{tab:Errors}, the error quantities are averaged over all text-descriptions of an instruction-setting (e.g. Length: Long, character: Casual, Style: Marketing-Message). The results underline our previous findings. Providing only an image as input information leaves more degrees of freedom for the description generation. While this leads to more diverse outputs, it also increases hallucination. Shorter descriptions seem to be more accurate than longer ones for image-only grounding. We observe a mean error of 1.048 for short descriptions and 1.215 for long descriptions, as the model has more room for hallucinations. On average, there are 1.13 labels described inaccurately per description. 

Only providing the ground-truth labels restricts the diversity of outputs but therefore noticeably limits hallucinations. The average error per description is reduced by a factor of 5.38.
If an image is passed together with the ground-truth label, the accuracy is lowered by about 28.6\% compared to label-only, but still surpasses the image-only accuracy by a factor of 4. The results are summarized in \Cref{tab:Errors}.
For these two cases, we observe that longer prompts are more accurate than shorter ones. For label-only inputs, short descriptions lead to 0.265 errors on average while long descriptions produce 0.15. For image and labels as the data source, short descriptions lead to a mean error of 0.355 per description and long ones to only 0.1725. 

\begin{table}[h]
\caption{Average net error counts for generated text-descriptions compared to the ground-truth annotations in GeoBiked. We compare different configurations of the system-prompt (Instruction Setting). The table reports the average number of categorical labels misclassified per generated description across different instruction settings and input configurations.}
\small
\renewcommand{\arraystretch}{1.3}
\centering
\begin{tabularx}{\columnwidth}{ZYYY}
\toprule
Instruction Setting & Image Only & Ground-Truth Label &   Image and Label \\
\midrule
Long; Casual; Marketing-message        &        0.72 &         0.19 &            0.20 \\
Long; Casual; Prompt-to-midjourney     &        1.05 &         0.20 &            0.27 \\
Long; Technical; Marketing-message     &        1.33 &         0.13 &            0.12 \\
Long; Technical; Prompt-to-midjourney  &        1.76 &         0.08 &            0.10 \\
Short; Casual; Marketing-message       &        0.77 &         0.36 &            0.40 \\
Short; Casual; Prompt-to-midjourney    &        1.03 &         0.31 &            0.34 \\
Short; Technical; Marketing-message    &        1.08 &         0.23 &            0.30 \\
Short; Technical; Prompt-to-midjourney &        1.31 &         0.16 &            0.38 \\
\midrule
\textbf{Mean} &  \textbf{1.13} & \textbf{0.21} & \textbf{0.27} \\
\bottomrule
\end{tabularx}
\label{tab:Errors}
\end{table}

\subsubsection{Implications}
For the task of automatically annotating technical images with feasible and diverse text descriptions, GPT-4o is an adequate choice. Due to the broad context understanding and the sufficient vision-language reasoning capabilities, domain-specific details are captured and described accurately. For the configuration of the input information, there exists a tradeoff between diversity of the generated text descriptions and their accuracy compared to ground-truth labels. While grounding the generation on labels leads to more accurate descriptions and reduces hallucinations, it restricts the diversity. For our example, pure label-grounding also leads to loss of information, as the labels contain significantly less details than the images. Providing an image paired with labels appears to reduce hallucinations while accurately capturing most visual features of the geometries. The diversity of text descriptions is noticeably reduced compared to only using an image as input information. Therefore, this configuration is somewhat of a tradeoff between diversity and accuracy. The configuration of instruction settings and input information ultimately depends on the desired application of the generated text descriptions. 

Although GPT-4o reliably adheres to the instructions stated in the system-prompt, we find that for large-scale automated annotation tasks, the system-prompt has to be evaluated and systematically optimized. The requirements and constraints for the model need to be explicitly included to avoid hallucinations. The system-prompt severely influences the quality of the outputs and their downstream usability. The same is true for employing GPT-4o as a label-classifier to infer categories from natural language texts. The model possesses sophisticated capabilities for this task that can be of benefit, as no use-case specific classifier has to be trained on the given problem. However, this does require careful prompt-engineering for the outputs to be in a structure that can be evaluated algorithmically. We find the DSPy-library of enormous help for such tasks \citep{khattab2023dspy}.

To process technical or CAD-like images with fine-grained visual details and generate adequate descriptions, large VLMs are necessary. In an ablation experiment, we instruct the significantly smaller Moondream model \citep{vikhyatkMoondream12024} with the same task and observe that the generated descriptions are very uniform. Unique descriptions are generated rarely. The descriptions that are unique show a low Levenshtein distance, as all of them follow the same sentence structure (see \Cref{fig:Levenshtein_Moondream}).

\section{Limitations and Future Work}
\label{sec:Sec4}
\subsection{GeoBiked Dataset} With the dataset, we aim to build a foundation to apply DGMs in engineering design processes. While real-world use-cases require data beyond bicycle images, comparing the performance of DGMs is an important first-step to foster the future development of the field. GeoBiked presents an opportunity to create benchmark-challenges for visual DGMs. There is however a need for evaluation metrics that capture the structural integrity and real-world feasiblity of the synthesized images. While there are approaches to better capture structural feasibility in images \citep{fan2024enhancingplausibilityevaluationgenerated, regenwetterStatisticalSimilarityRethinking2023}, this largely remains an open challenge. Additionally, the dataset currently does not include any explicit features to evaluate the performance of the bicycle designs. Adding this is another step towards improving applicability in engineering design domains.

\subsection{DGM-driven Engineering Design Applications}
\label{subsec: DGM-Application}
Our objective with GeoBiked is to support and inspire the application of DGMs in engineering design tasks. By providing a dataset with both geometric and semantic annotations, we aim to offer a resource that can be used for initial experimentation and exploration, accelerating the adoption of DGMs in real-world engineering design use cases. In this section, we briefly discuss potential applications of GeoBiked and describe some examples.

\textbf{Model \& Architecture verification.} GeoBiked can serve as a resource for evaluating the applicability of different DGM architectures in engineering design tasks. By training multiple DGM-types on the dataset, practitioners can gain insights into how well different models handle structured technical image data. This kind of experimentation helps to assess computational requirements, dataset suitability and model limitations. We provide an exploratory example of this in \Cref{Sec:DGM_Training}.

\textbf{Conditioning.} For engineering design applications, the possibility to condition the DGM on relevant modalities is a fundamental requirement. Developing custom conditioning mechanisms either requires training data, for example to train a ControlNet adapter \citep{zhangAddingConditionalControl2023}, or test data to optimize and verify the method. GeoBiked can help with these tasks. In a different work, the dataset is used to enable a training-free architecture for visualization of engineering design images in realistic scenes \citep{MuellerInsertDiffusion2024}.
\Cref{sec:TextGen}.

\textbf{Design Optimization.} A promising direction for future work involves exploring the use of GeoBiked for design optimization tasks, such as performance-driven shape generation or constraint-aware synthesis. The dataset’s geometric reference points and semantic labels could support conditional generative models, where design constraints are incorporated as inputs for more targeted generation. However, for effective design optimization, the dataset would require the inclusion of performance-related parameters that could serve as objective functions for optimization tasks. Expanding the dataset with such information remains an important step for enabling data-driven design optimization workflows.

\subsection{Geometric Feature Detection with Diffusion Hyperfeatures.} While we show that Diffusion Hyperfeatures can be used to draw semantic correspondences in technical images and label the data in a few-shot fashion, the technique most likely requires refinement for more complex structures. Possible directions are retraining or finetuning the aggregation network for use-case specific problems or modifying the similarity computation to compare Hyperfeatures of patches rather than singular points. Overall, we find that is method is an exiting approach worth exploring for other problems such as object consistency and similarity in image and 3D generation.

\subsection{Automatic Generation of Text Descriptions with VLMs.}
The dependence on GPT-4o, as a model only accessible through the API without open-source weights is a drawback. While this solution is easy-to-use, we encourage investigating open-source alternatives such as LLaVA \citep{liuVisualInstructionTuning2023, liuImprovedBaselinesVisual2024} for this task. Vision-language understanding capabilities are likely to improve rapidly in open-source alternatives.

\section{Conclusion}
\label{sec:Sec5}
 We illustrate the potential of utilizing large-scale pretrained generative models to address the challenges of domain-specific tasks in engineering design. The GeoBiked dataset serves as a resource for baseline investigations into model feasibility, data requirements, and benchmarking in engineering design applications, providing a foundation for exploration of Deep Generative Models (DGMs). Additionally, we have demonstrated that off-the-shelf visual foundation models, when used effectively and guided correctly, can automate data labeling and annotation tasks, significantly lowering the barriers to entry for DGMs in technical fields. 
 
 Diffusion models for image generation inherit spatial and semantic understanding that can be used to draw geometric correspondences in structural images by consolidating image features into an interpretable Hyperfeature map. The prediction of geometric reference points in unseen images of technical objects is significantly improved if multiple examples, which are showing different styles of objects are used as reference. Large vision-language models are applicable to automatically generate text descriptions of technical images. The accuracy and creativity of the generated descriptions depend on careful prompting and the provided input information.
 Overall, our findings aim to facilitate the broader adoption of AI-driven approaches in engineering design, streamlining processes and expanding creative possibilities in the field.

\bibliographystyle{plainnat}  
\bibliography{GeoBIKED}

\pagebreak

\appendix
\section*{Appendix}
\label{sec:Appendix}

\section{Dataset Distribution}
\begin{figure}[h]
\centering
\includegraphics[width=\linewidth]{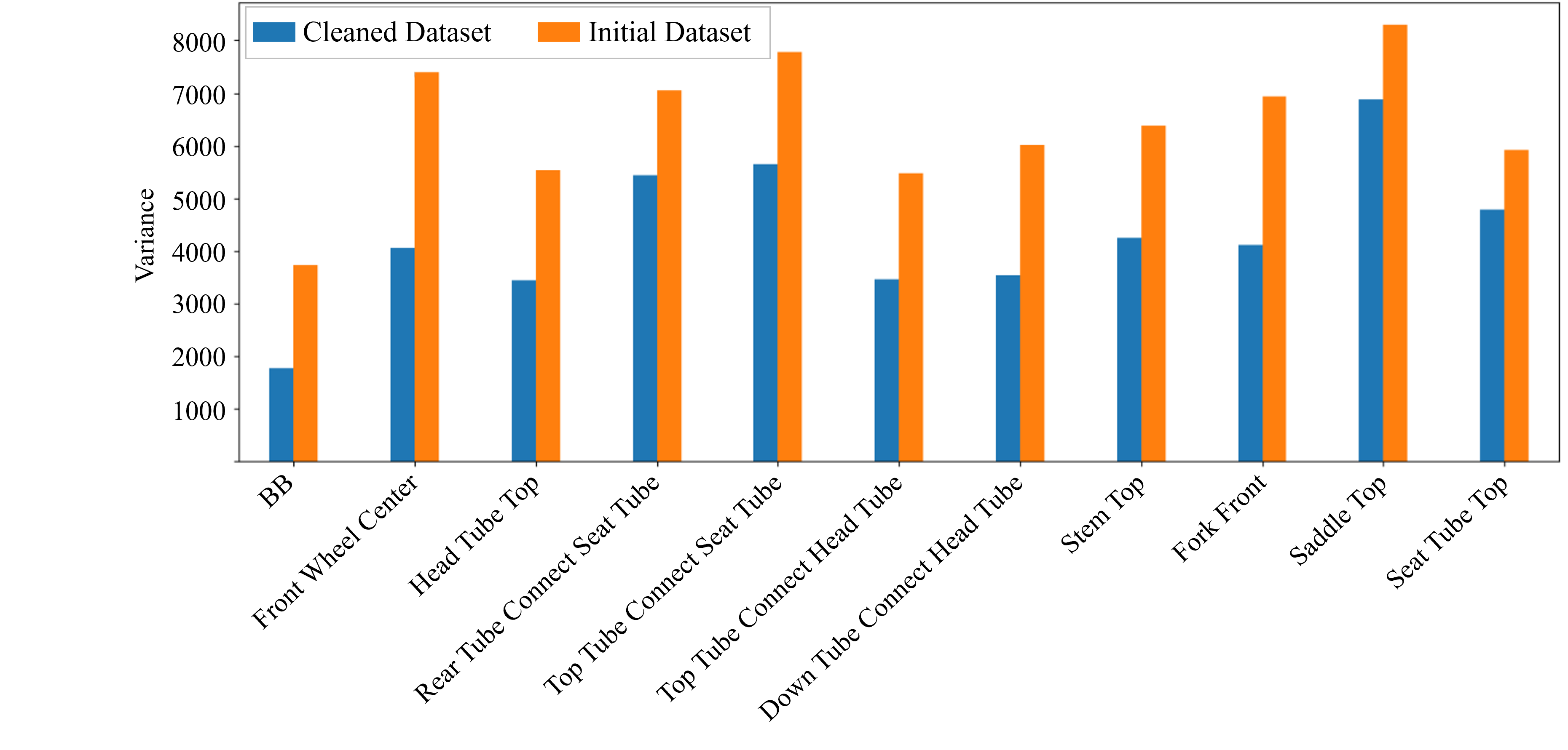}
\caption{Variance of geometric bicycle layouts before (orange) and after (blue) the filtering of the dataset. The blue bars represent GeoBiked, while the orange bars represent Biked \citep{regenwetterBIKEDDatasetMachine2021}.}
\label{fig:Variance_Reduction}
\end{figure}

\FloatBarrier
\section{Training visual DGMs on GeoBiked}
\label{Sec:DGM_Training}
To provide a practical demonstration of how the GeoBiked dataset can be utilized for training DGMs, we conduct a proof-of-concept experiment using a selection of common generative architectures. Our aim is not to establish benchmark performance improvements but rather to illustrate the dataset’s compatibility with various DGM approaches and highlight considerations for model selection and training duration, reflecting challenges practitioners might encounter when working with domain-specific datasets.

\begin{figure}[h]
\centering
\includegraphics[width=\linewidth]{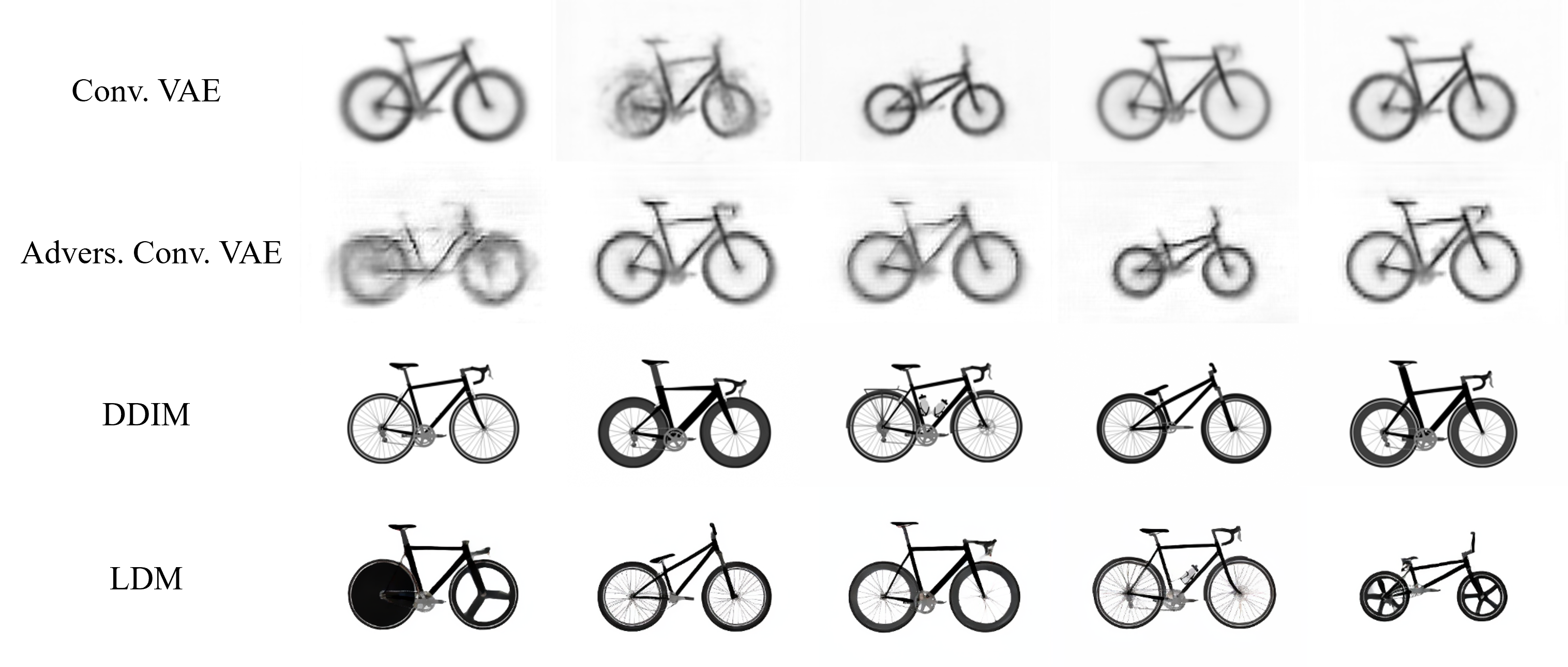}
\caption{Generated samples with DGMs trained on GeoBiked. We show visually feasible results that reflect a range of features and diversity to illustrate the types of outputs the models can generate.}
\label{fig:ApplicationExamps}
\end{figure}

In this exploratory example, we train four different DGMs on the GeoBiked dataset: a convolutional variational autoencoder (CVAE) \citep{kingmaAutoEncodingVariationalBayes2013, sohnLearningStructuredOutput2015, rombachHighResolutionImageSynthesis2022}, a variant of the CVAE with adversarial training (Adversarial CVAE) \citep{BlattmanniPoke2021}, a Denoising Diffusion Implicit Model (DDIM) \citep{hoDenoisingDiffusionProbabilistic2020, songDenoisingDiffusionImplicit2022}, and a Latent Diffusion Model (LDM) \citep{rombachHighResolutionImageSynthesis2022}. For the DDIM, we use the implementation provided by \citep{fanNoiseSchedulingGenerating2023} and for the LDM we use the implementation from \citep{FisherPlantLDM2022}. All models were trained from scratch on an NVIDIA RTX A6000 Ada GPU. The VAEs were trained using 10GB of VRAM, while the LDM required 17.5GB and the DDIM 20GB. All models were trained purely on image data without additional conditioning information. Our goal is to show   rather than optimizing for absolute generative performance. The goal of this experiment is not to optimize generative model performance but to show how these models handle structural image data and explore variations in visual quality across different architectures. We therefore aim to provide a high-level demonstration of how the GeoBiked dataset can be used for practical comparisons like the evaluation of training costs (in terms of computational requirements) and output quality. 
A brief summary of the results is provided in \Cref{tab:ModelFeasibility} and \Cref{fig:ApplicationExamps} shows some qualitative examples. 

While both diffusion-based models are able to reproduce the structural image data, the VAE-architecture struggles to synthesize clean images. Even with the introduction of an adversarial loss, the results that can be synthesized remain very blurry. In terms of compute requirements, the VAE is much cheaper but its capabilities are not sufficient for detailed images with fine-grained structures. Even though there exist Autoencoder-based approaches for such image data \citep{fanAdversarialLatentAutoencoder2023}, we observe that off-the-shelf diffusion models handle this data much better. Comparing both diffusion-models, we observe that in both quantitative (FID-score \citep{szegedyGoingDeeperConvolutions2014}) and qualitative measures (\Cref{fig:ApplicationExamps}), the LDM is outperformed by the DDIM. The DDIM, although significantly more expensive in training, reproduces the fine-grained structures better. Due to the images being encoded into a latent representation before learning the diffusion model, the LDM loses some structural and geometric information and the results are more blurred.

\begin{table}[h]
\caption{Model feasibility study. We compare four different DGM-architectures to synthesize the GeoBiked image distribution. All models were trained on an RTX A6000 ADA GPU with 48GB of VRAM capacity. Due to limited capacity, both VAE models are trained on 10GB of VRAM, while the LDM is trained on 17.5GB and the DDIM on 20GB. The training durations are normalized for 20GB VRAM. While VAE-architectures require significantly less compute, they do not allow for the generation of high quality outputs with  structural details. The DDIM shows the best FID-score, but requires the most computational resources.}
\centering
\begin{tabularx}{\columnwidth}{YYY}
\toprule
{Architecture} & {FID $\downarrow$} & {Normalized Training Duration (hr)}\\
\midrule
\textit{CVAE} & 123.85 & 0.23 \\
\textit{Adversarial CVAE} & 166.98 & 0.25 \\
\textit{DDIM} & 12.03 & 68.05 \\
\textit{LDM} & 20.69 & 30.18 \\
\bottomrule
\end{tabularx}
\label{tab:ModelFeasibility}
\end{table}

\FloatBarrier

\section{Geometric Feature Detection with Diffusion Hyperfeatures - Experiment Details}
\begin{figure}[h]
\centering
\includegraphics[width = \linewidth]{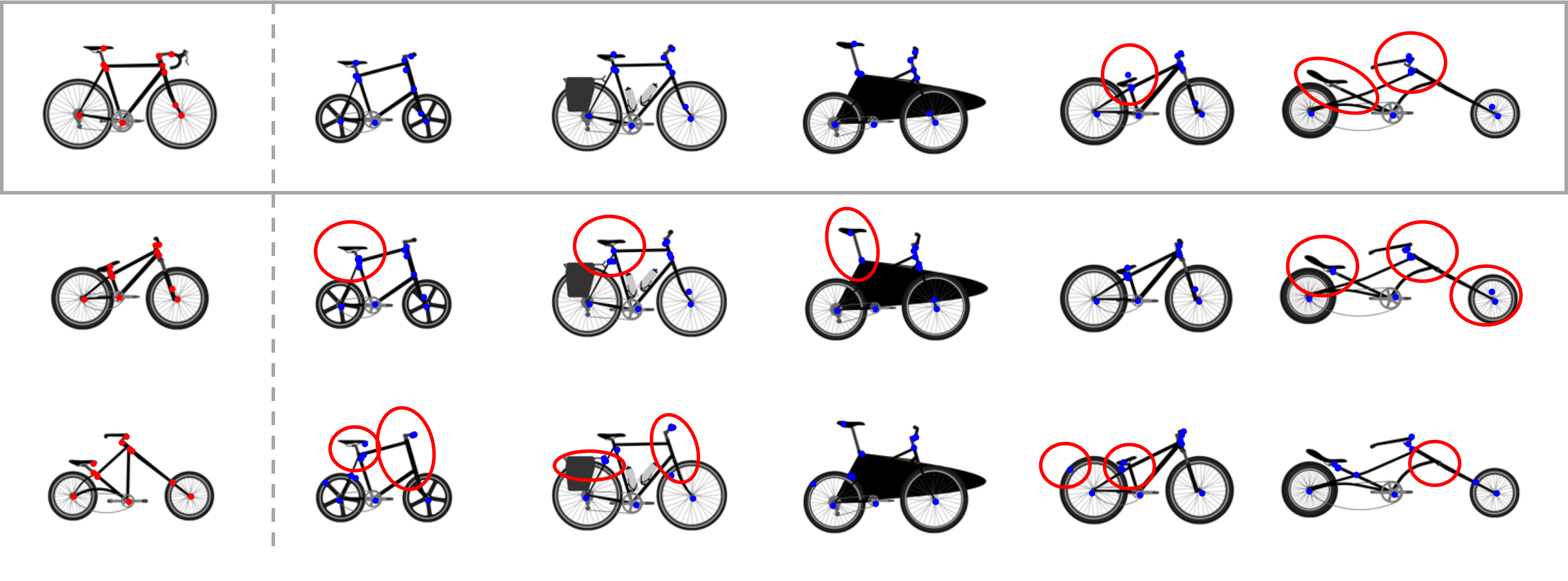}
\caption{Qualitative Comparison of Error Patterns for a single source image. Top row shows the source image with the highest predicition accuracy. Left column shows the input images with the manually annotated reference points, right columns the automatically annotated samples. Red circles mark areas of uncertainty in the annotation.}
\label{fig:Error_Patterns_SSI}
\end{figure}

When using a single source image, the accuracy of the point detection heavily depends on the source images. An average sample that shows a bicylce of very common style and high similarity with many samples in the dataset produces good mean prediction accuracy (\Cref{tab:SingleSourceImg}). Outliers however are not predicted well (see \Cref{fig:Error_Patterns_SSI}, top row). Typical error patterns include uncertainty in the detection of the saddle top and tube intersections. When other bicycle styles are used as source images, the prediction accuracy deteriorates significantly. Using a BMW-style as source, the saddle is not detected for any other style than BMW (\Cref{fig:Error_Patterns_SSI}, middle row). When an uncommon style is used as source, only the wheel centers are accurately detected. All other points are prone to significant inaccuracies (\Cref{fig:Error_Patterns_SSI}, bottom row).

\begin{table}[h]
\caption{Experiment results for a single source image. The metrics are computed as a pixel-wise distance between prediction and ground truth and averaged over all 12 geometry-points. The duration measures the processing time of the subset of 150 samples. Image indices from left to right are 1, 2, 31 and 74.}
\centering
\begin{tabularx}{\columnwidth}{cYYYY}
\toprule
{Source}
& 
\begin{minipage}{.2\textwidth}
    \centering
    \includegraphics[scale = 0.4]{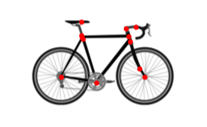}	
\end{minipage} 
&
\begin{minipage}{.2\textwidth}
    \centering
    \includegraphics[scale = 0.4]{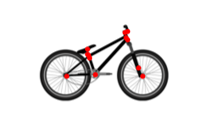}	
\end{minipage}
&
\begin{minipage}{.2\textwidth}
    \centering
    \includegraphics[scale = 0.4]{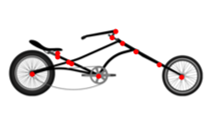}	
\end{minipage}
&
\begin{minipage}{.2\textwidth}
    \centering
    \includegraphics[scale = 0.4]{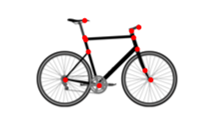}	
\end{minipage}\\
\midrule
\textit{MAE} & 2.429 & 3.385 & 12.274 & 2.710 \\
\textit{MSE} & 33.158 & 41.635 & 482.969 & 41.272 \\
\bottomrule
\end{tabularx}
\label{tab:SingleSourceImg}
\end{table}

\begin{figure}[h]
\centering
\includegraphics[width = \linewidth]{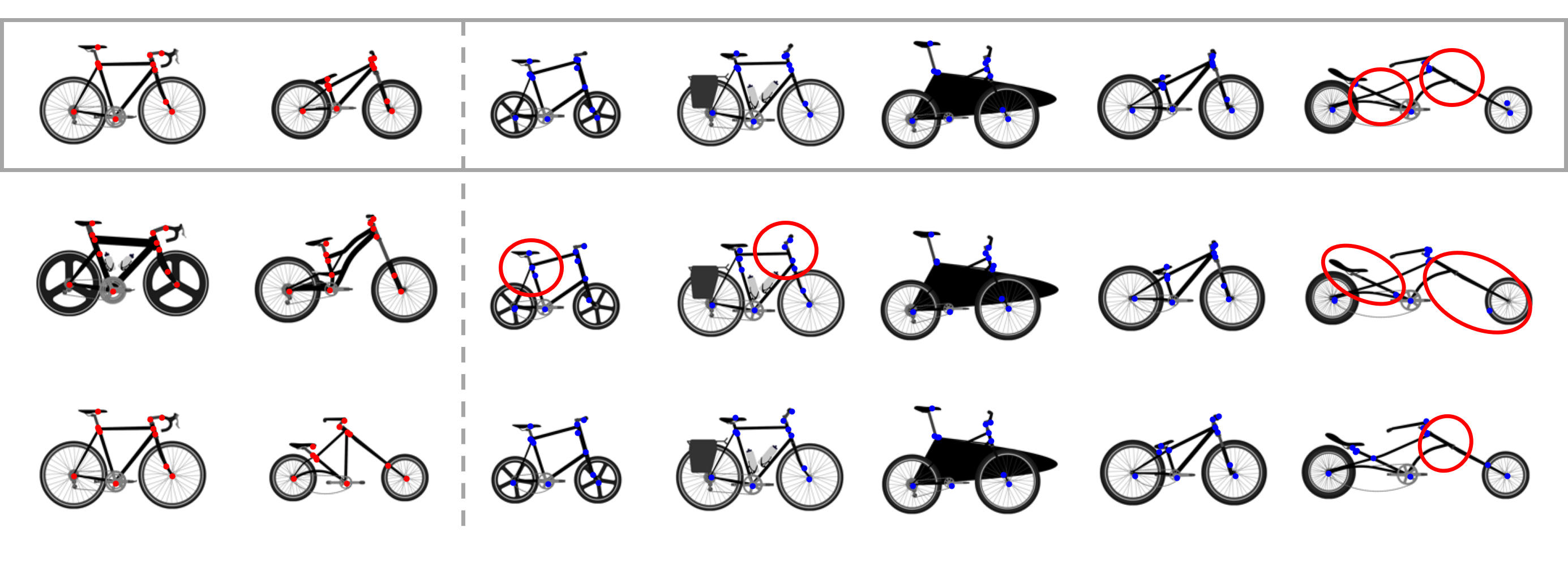}
\caption{Qualitative Comparison of Error Patterns using two source images. Top row shows the combination of source images with the best prediction accuracy. Left column shows the input images with the manually annotated reference points, right columns the automatically annotated samples. Red circles mark areas of uncertainty in the annotation.}
\label{fig:Error_Patterns_2SI}
\end{figure}

\begin{figure}[h]
\centering
\includegraphics[width = \linewidth]{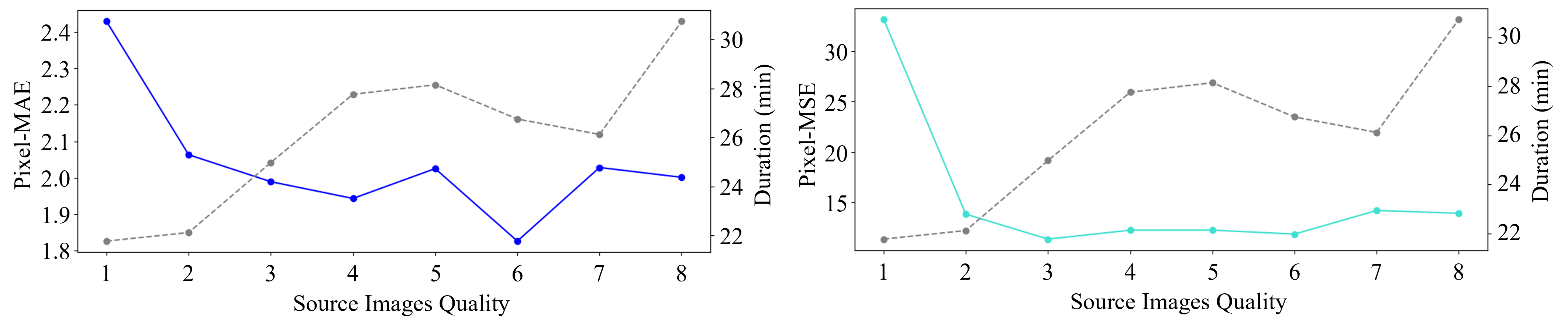}
\caption{Pixel-wise MAE, MSE and Processing Duration over the number of annotated source images.}
\label{fig:Error_over_Quant}
\end{figure}

\FloatBarrier

\section{Automatic Generation of Text Descriptions with Vision-Language Models}
\subsection{GPT-4o System-Prompt}
\label{sec:Code}
\begin{lstlisting}
def construct_prompt(
        length: str,
        character: str,
        style: str,
        mode: Literal["im-only", "txt-grounded", "im-txt-grounded"] = "im-only",
) -> str:
    length_map = {
        "short": "between 5 and 10 words",
        "medium": "between 10 and 20 words",
        "long": "between 20 and 40 words",
    }
    mode_task_map = {
        "im-only": "images",
        "txt-grounded": "technical data about bicycles",
        "im-txt-grounded": "images and technical data about bicycles contained in them",
    }

    image_wrapper = 'Images will be wrapped between <image i></image i> tags.\n' if 'im' in mode else ''
    bike_wrapper = 'Bike data will be wrapped between <data i></data i> tags.\n' if 'txt' in mode else ''
    description_usage = (
        'You do not have to include all, or any, '
        'of the bike data in the description if it does not fit the style or character. '
        'It is important that the description fits the constraints mentioned above.\n'
        if "txt" in mode else ''
    )

    prompt = (
        "Your task is to create descriptions of bicycles based on {0}. ".format(mode_task_map[mode])
        + "Each description should fulfill the following constraints: \n"
        + "- The length of the provided description should be {0}. \n".format(length_map[length])
        + "- The descriptions should be {0}. \n".format(character)
        + "- The descriptions should be in the style of a {0}. \n".format(style)
        + "{0}".format(image_wrapper)
        + "{0}".format(bike_wrapper)
        + "Wrap the resulting bike description in <description i></descriptions i> tags.\n"
        + "There should be *no* newlines in the descriptions.\n"
        + "{0}".format(description_usage)
        + "The descriptions should be very diverse within the given constraints."
    )

    return prompt
\end{lstlisting}
\pagebreak
\FloatBarrier

\subsection{Examples of Generated Description}

\begin{table}[h]
\caption{Examples of generated text descriptions for different bicycles from the GeoBiked dataset. Descriptions are generated with different configurations of the system prompt and for all three input information combinations.}
\includegraphics[width = \linewidth]{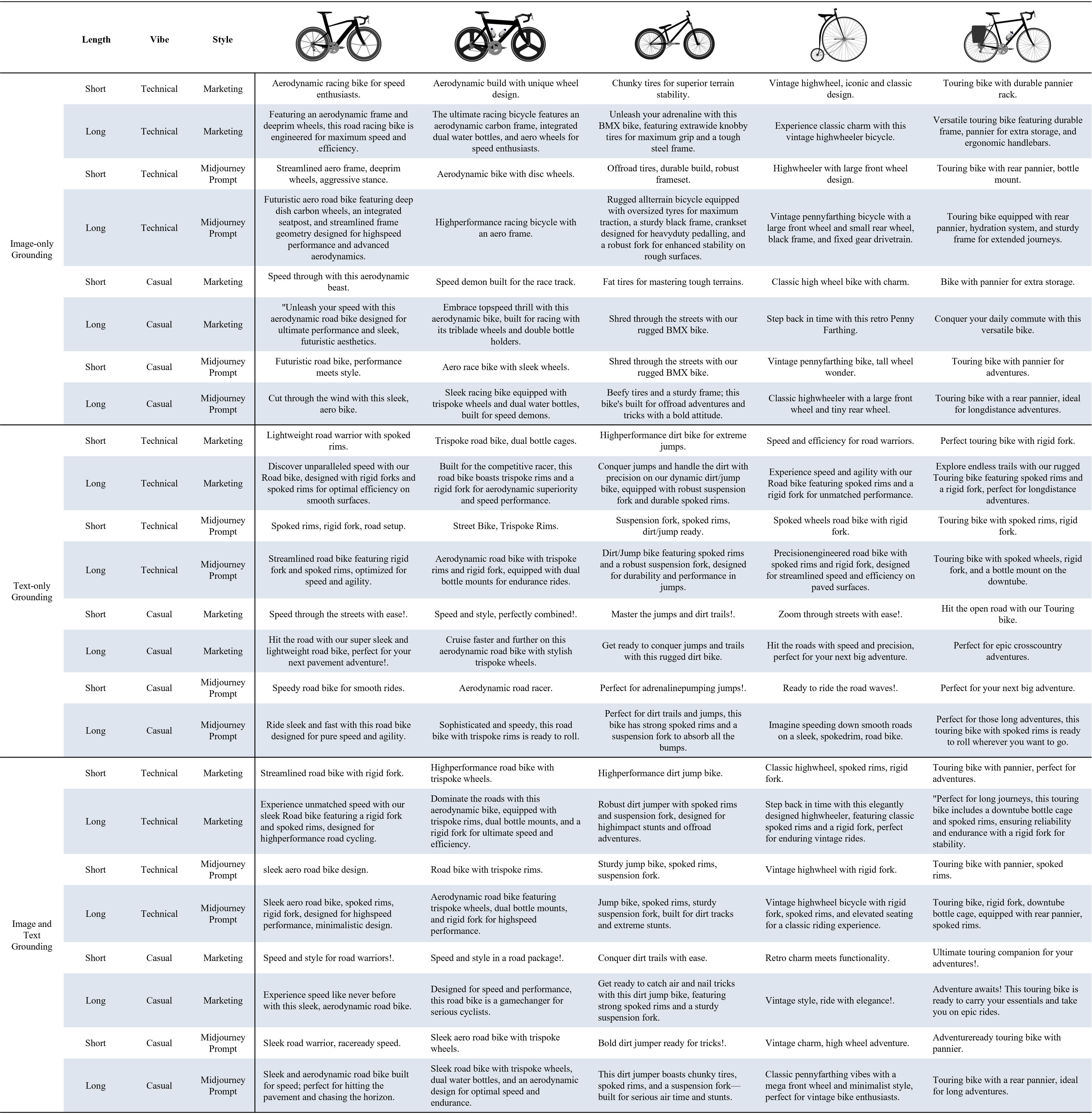}
\label{tab:Example_Descriptions}
\end{table}

\pagebreak
\subsection{Gpt-4o vs. Moondream}
\begin{figure}[h]
\centering
\includegraphics[width = \linewidth]{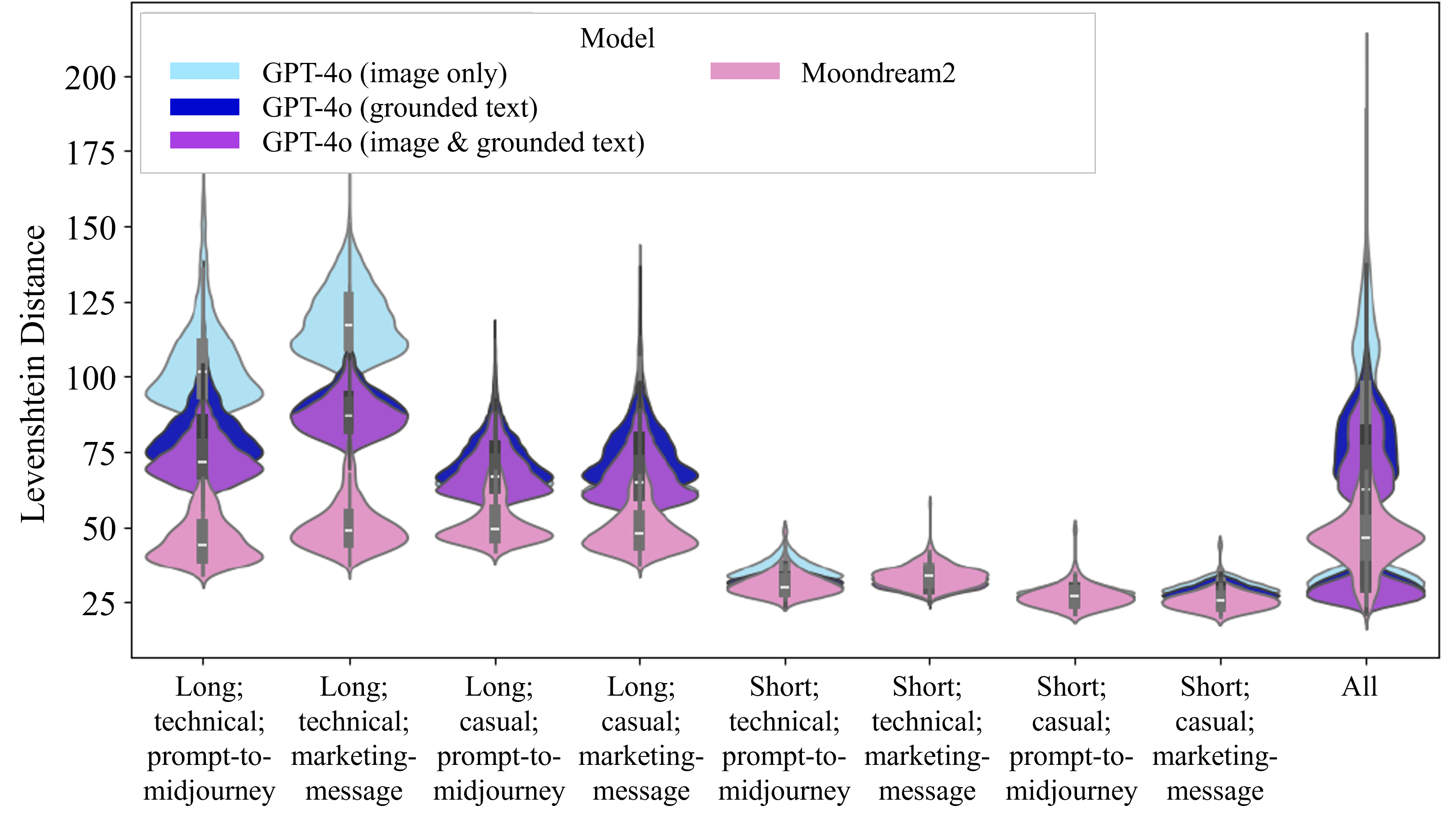}
\caption{Levenshtein distance of unique outputs generated by GPT-4o configurations compared to Moondream. The descriptions generated by Moondream are significantly less diverse.}
\label{fig:Levenshtein_Moondream}
\end{figure}

\end{document}